\pdfoutput=1

\documentclass[11pt]{article}

\usepackage{style/acl_arr}
\usepackage{times}
\usepackage{latexsym}

\usepackage[T1]{fontenc}

\usepackage[utf8]{inputenc}
\usepackage{microtype}
\usepackage{booktabs} 
\usepackage{siunitx}  
\makeatletter

\renewcommand{\paragraph}{%
  \@startsection{paragraph}{4}%
  {\z@}{1ex \@plus 1ex \@minus .2ex}{-1em}%
  {\normalfont\normalsize\bfseries}%
}
\makeatother
\usepackage{makecell}

\newcommand{\advscore}{\abr{AdvScore}}
\newcommand{\advqa}{\abr{AdvQA}}
\newcommand{\gpt}{\abr{gpt}}
\newcommand{\irt}{\abr{irt}}
\newcommand{\twoplirt}{\abr{{\small{2}}pl-irt}}
\newcommand{\iif}{\abr{iif}}

\newcommand{\team}{group}
\newcommand{\subj}[0]{subject}
\usepackage{float}
\usepackage{makecell} 
\usepackage{algorithm}
\usepackage[noend]{algpseudocode}
\usepackage{varwidth}
\usepackage{multirow}
\usepackage{tabularx}
\usepackage{subcaption}
\usepackage{amssymb}
\usepackage{amsmath}
\usepackage{amsthm}
\usepackage{amsfonts}
\usepackage{bbm}
\usepackage{bm}
\usepackage{booktabs}
\usepackage{graphicx}
\usepackage{multirow}
\usepackage{multicol}
\usepackage{enumitem}
\usepackage{mdwlist}
\usepackage{csquotes}
\usepackage{multirow}
\include{cv-commands}
\setlist{nosep}
\usepackage{soul}
\usepackage{xcolor}
\usepackage{tcolorbox}

\title{Is your benchmark \textit{truly} adversarial? \\
\advscore{}: Evaluating Human-Grounded Adversarialness}

\author{
    Yoo Yeon Sung$^{1}$, Maharshi Gor$^{1}$, Eve Fleisig$^{2}$, Ishani Mondal$^{1}$, Jordan Boyd-Graber$^{1}$ \\[1em]
    $^{1}$University of Maryland 
    $^{2}$UC Berkeley \\[1em]
}

\date{}

\newif\ifcomment\commenttrue

\usepackage[a-1b]{pdfx}

\usepackage{framed}
\usepackage{mdwlist}
\usepackage{siunitx}
\usepackage{latexsym}
\usepackage{colortbl}
\usepackage{xcolor}
\usepackage{nicefrac}
\usepackage{booktabs}
\usepackage{fnpct}
\usepackage{amsfonts}
\usepackage[T1]{fontenc}
\usepackage{bold-extra}
\usepackage{amsmath}
\usepackage{amssymb}
\usepackage{bm}
\usepackage{graphicx}
\usepackage{mathtools}
\usepackage{microtype}
\usepackage{multirow}
\usepackage{multicol}
\usepackage{xpatch}
\usepackage{latexsym,comment}
\usepackage[normalem]{ulem}

\newcommand*{\missingreference}{{\Huge \colorbox{red}{?reference?}}}
\newcommand*{\missingcitation}{{\Huge \colorbox{red}{?citation?}}}

\makeatletter
\xpatchcmd{\@setref}{\bfseries}{\missingreference}{}{}
\def\@citex[#1]#2{\leavevmode
    \let\@citea\@empty
    \@cite{\@for\@citeb:=#2\do
        {\@citea\def\@citea{,\penalty\@m\ }%
            \edef\@citeb{\expandafter\@firstofone\@citeb\@empty}%
            \if@filesw\immediate\write\@auxout{\string\citation{\@citeb}}\fi
            \@ifundefined{b@\@citeb}{\hbox{\reset@font\missingcitation}%
                \G@refundefinedtrue
                \@latex@warning
                {Citation `\@citeb' on page \thepage \space undefined}}%
            {\@cite@ofmt{\csname b@\@citeb\endcsname}}}}{#1}}
\makeatother

\newcommand{\gem}[1]{\mbox{\textsc{gem}}}
\newcommand{\abr}[1]{\textsc{#1}}

\newcommand{\explain}[2]{\underbrace{#2}_{\mbox{\footnotesize{#1}}}}

\newcommand{\g}{\, | \,}

\DeclareUnicodeCharacter{03B2}{ }
\DeclareUnicodeCharacter{2212}{ }

\newcommand{\hidetext}[1]{}
\newcommand{\ignore}[1]{}

\ifcomment
    \newcommand{\pinaforecomment}[3]{\colorbox{#1}{\parbox{.8\linewidth}{#2: #3}}}

    \newcommand{\prtodo}[1]{\pinaforecomment{lightblue}{pr}{#1}}
    \newcommand{\prtodoi}[1]{\pinaforecomment{lightblue}{pr}{#1}}
\else
    \newcommand{\pinaforecomment}[3]{}
    \newcommand{\prtodo}[1]{}
    \newcommand{\prtodoi}[1]{}
\fi

\newcommand{\jbgcomment}[1]{\pinaforecomment{red}{JBG}{#1}}

\newcommand{\smallurl}[1]{ \begin{tiny}\url{#1}\end{tiny}}

\definecolor{lightblue}{HTML}{3cc7ea}
\definecolor{CUgold}{HTML}{CFB87C}
\definecolor{grey}{rgb}{0.95,0.95,0.95}
\definecolor{ceil}{rgb}{0.57, 0.63, 0.81}
\definecolor{UMDred}{HTML}{ed1c24}
\definecolor{UMDyellow}{HTML}{ffc20e}

\newcommand{\ai}[0]{\textsc{ai}}

\newcommand{\qa}[0]{\abr{qa}}

\newcommand{\hitl}[0]{\abr{hitl}}

\newcommand{\squad}{\textsc{sq}{\small u}\textsc{ad}}
\newcommand{\roberta}{\textsc{r}{\small o}\textsc{b}{\small erta}}

\newcommand{\farmreader}{\textsc{f}{\small arm}\textsc{R}{\small eader}}

\newcommand{\distilbert}{\textsc{d}{\small istil}\textsc{b}{\small ert}}

\usepackage{pifont}
\usepackage{arydshln}
\usepackage[normalem]{ulem}

\usepackage{tikz}
\usetikzlibrary{patterns}

\newtcbox{\entoure}[1][red]{on line, arc=3pt,colframe=#1!50!black, 
before upper={\rule[-3pt]{0pt}{10pt}},boxrule=1pt,
boxsep=0pt,left=2pt,right=2pt,top=1pt,bottom=.5pt}

\graphicspath{ {figures/}{auto_fig/} }

\begin{document}
\maketitle
\begin{abstract}
\newif\ifcomments
\commentstrue 
\ifcomments
    \newcommand{\evef}[1]{{\protect\color{orange}{[EF: #1]}}}
    \newcommand{\jess}[1]{{\protect\color{blue}{[JD: #1]}}}
\else
    \newcommand{\evef}[1]{}
    \newcommand{\jess}[1]{}
\fi

Adversarial datasets should validate AI robustness by providing samples on which humans perform well, but models do not. However, as models evolve, datasets can become
obsolete.
Measuring whether a dataset remains adversarial is hindered by the lack of a standardized metric for measuring adversarialness.
We propose \advscore{}, a human-grounded evaluation
metric that assesses a dataset's adversarialness
by capturing models' and humans' varying abilities, while also
identifying poor examples.
We then use \advscore{} to motivate a new dataset creation pipeline
for realistic and high-quality adversarial samples, enabling us to
collect an adversarial question answering (\qa{}) dataset, \advqa{}.
We apply \advscore{} using 9,347 human responses and ten language
models' predictions to track model improvement over five years
(2020--2024).
\advscore{} thus provides guidance for achieving robustness comparable with human capabilities. Furthermore, it helps determine to what extent adversarial datasets continue to pose challenges, ensuring that, rather than reflecting outdated or overly artificial difficulties, they effectively test model capabilities.\footnote{Code and data available here: \url{github.com/yysung/Advscore}}

\end{abstract}
\section{Introduction: Evaluating Adversarial Datasets Requires Human Answers} \label{intro}
\begin{figure*}[!t] 
    \centering
    \includegraphics[width=\linewidth]{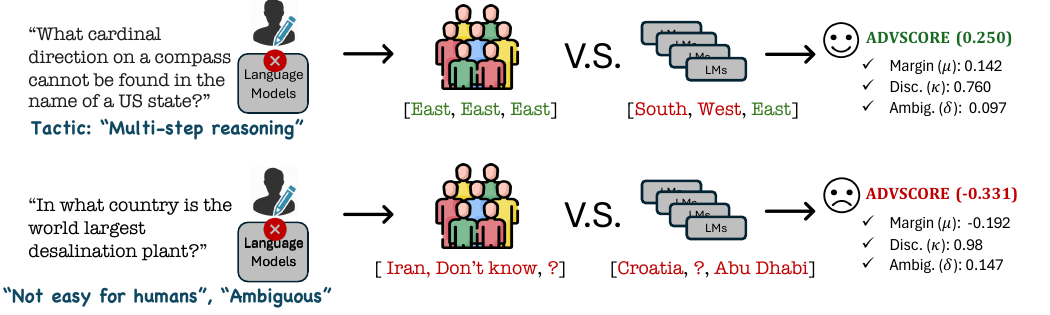}
    \caption{\advscore{} diagnoses when a question is adversarial (top) and difficult for computers to answer for other reasons (bottom).  After collecting candidate questions, we ask humans and computers to answer the questions.  The top question (from \advqa{}) has a higher \advscore{} because it is specific, adversarial, discriminative, high-quality, and
      realistic. In contrast, the bottom question is 
      ambiguous (e.g., none of humans or models
      correctly answered due to its ambiguity), which is confirmed by its 
      low \advscore{}.}
    \label{fig:adversarial_creation_process}
\end{figure*}


As language models attain near-perfect performance on existing
benchmarks, there is an increasing demand for unexpected and
challenging tasks to evaluate them.
\textit{Adversarial datasets} contain examples that cause models to generate
harmful~\citep{perez2022red}, unsafe~\citep{quaye2024adversarial}, or
incorrect~\citep{goodfellow2015explaining} responses.
An ideal adversarial example should be much easier for a human to answer correctly than for a model on realistic tasks~\citep{ilyas2019adversarial, tsipras2019robustness,
  engstrom2019adversarial, biggio2012poisoning}.
%
However, as models improve, these adversarial datasets can become
outdated~\citep{kiela-etal-2021-dynabench}---what was hard for a model in 2020 can become trivial in five
years---requiring periodic updates~\citep{pmlr-v97-recht19a, bowman-dahl-2021-will}. On the other
hand, it is difficult to recognize at what point have these
adversarial datasets outlived their usefulness systematically, nor is
there an established metric to measure which datasets best captures the
gap between human and model ability.
%

To fill this gap, we formulate \textbf{\advscore{}}
(\S~\ref{section:metric}). This metric measures two critical aspects:
\textbf{(i) adversarialness}, which captures the performance gap
between models and humans, while penalizing ``ill-posed'' examples
(i.e., ambiguity), and \textbf{(ii) discriminability}---how
effectively can a dataset rank models by their abilities.

Measuring whether a dataset is truly adversarial requires human
answers; thus, \advscore{} builds on item response
theory~\citep[\irt{}]{lalor2016building}, a framework widely used in
psychometrics and educational testing.  It captures the diversity of
human and model abilities and identifies poor examples
(\S~\ref{sec:irt}).
%
\advscore{} is the first metric that evaluates an example's
``adversarialness'' grounded in human abilities: it can measure
whether the dataset's adversarial challenge becomes weaker or stronger
as language models improve.

We apply \advscore{} to motivate authors to contribute to a new
human-in-the-loop \abr{hitl} benchmark of adversarial questions,
\advqa{}.
\advqa{}'s creation pipeline (Figure~\ref{fig:adversarial_creation_process}) produces
\textit{high-quality} and \textit{realistic} questions that are
adversarial. Moreover, \advscore{} helps make \advqa{} discriminative,
ensuring that the captured adversarialness reflects the varying skills
of humans and models.

\advqa{} exhibits the least decline in  adversarialness over 
recent years compared to other adversarial benchmarks
(\S~\ref{sec:comparison}).
This minimal, but meaningful decline in \advqa{} reveals that current
models (e.g., \gpt{}4) continue to struggle with tasks requiring
\textit{commonsense reasoning} and \textit{multistep reasoning} and on
topics such as \textit{Lifestyle}~(\S~\ref{sec:analysis}), which are
likely tied to real-world challenges.

We conclude with an analysis of how model have improved improve over
the years since researchers began releasing adversarial datasets and
how that can inform the development of future adversarial datasets
(\S~\ref{sec:comparison}).

\section{Preliminaries of \advscore: \irt{}}\label{sec:irt}

Prior metrics for evaluating adversarial question generation
strategies, such as attack success rate~\citep{uesato2018adversarial},
distributional similarity~\citep{dathathri2019plug}, and proximity
measurement~\citep{ross2021explaining} assess algorithmic
adversarialness without human validation.
In contrast, we identify adversarial examples that pose realistic
challenges aligned with \emph{human} skills, not just pathological cases that
break models.
This requires evaluating how well the examples align with varying
levels of human performance, particularly where models fall short,
while ensuring that the examples are unambiguous.
To capture this, we adopt item response theory (\irt{}), which models
the interactions between subjects' skills---in the QA setting, the subject
answering the question could be either a human or a model---and
example difficulty.
This framework, widely used in psychometrics and educational
testing~\citep{lord1968statistical}, provides insights beyond
accuracy: it can diagnose question quality as well as skilled
subjects.

\paragraph{\twoplirt{}}
In question answering~(\qa{}) tasks, \irt{} models the probability
that a subject correctly answers a question based on their skill and
question difficulties.
\twoplirt{} (Eq.~\ref{eq:2pl}) models the probability of getting a
question correct as a function of subject \emph{skill}~$\beta_i$ and
question \emph{difficulty}~$\theta_j$:
\begin{equation} \label{eq:2pl}
p(r_{ij} = 1 \g \beta_i, \theta_j, \gamma_j) = \sigma(\gamma_j(\explain{skill gap}{\beta_i-\theta_j})),
\end{equation}  
where $\sigma$ is the sigmoid function~\citep{baker2004item}. 
The skill gap, $(\beta_i-\theta_j)$, is the difference between the subject $i$'s skill and question $j$. When a subject's skill is \textit{equal} to the question's difficulty $(\beta_i = \theta_j)$, they have a 50\% probability of answering it correctly. Thus, an agent with skill equal to or greater than the question's difficulty level has at least a 50\% chance of answering correctly.

%
The final latent variable is the question
\emph{discriminability}~$\gamma_j$ which models how sensitive this
probability is to changes in skill gap.\footnote{Perfect discriminability means that any
\subj{}s with a positive skill gap will answer the question
correctly~\citep{martinez2019item} but negative skill gap will never
answer the question correctly.}
This encodes how strongly
the question rewards the skill being higher or lower than the
difficulty level. The objective of \irt{} is to
estimate the parameters that maximize the correctness probability
$p(r_{ij})$.\footnote{Implementation details in
Appendix~\ref{app:irt_detail}.}

\paragraph{Advantages of \irt{} over question success rate}\label{subsec:qsr}
While question success rate (\abr{qsr})---the percentage of subjects
answering a question correctly---may seem like a reliable measure of
difficulty, it can be misleading. A good yet difficult question and an
easy yet poorly written question could yield the same \abr{qsr}, obscuring the true measure of difficulty.

In contrast, \irt{} evaluates subject responses. 
%
%
Not only does \irt{} consider the number of humans who answer a question correctly, but it also accounts for \textit{who} answer \textit{which questions}. If the probability of answering a question correctly increases with subject skill, this relationship will naturally correlate with skill~$\beta_i$ and question discriminability~$\gamma_j$.
The model can confidently assign higher
probabilities for these questions, while questions that are answered
correctly by luck---rather than skill---will have estimated probabilities closer to 0.5, reflecting their lower discriminability.

\newcommand{\namedq}[1]{${q_{\text{#1}}}$}
Consider three questions:
\namedq{ambig} (ambiguous question: ``What is a capital of Georgia?''
Answer: [\underline{Atlanta} or \underline{Tbilisi}]), \namedq{hard}
(hard but well-formed question: ``Who founded Tbilisi?''), and
\namedq{easy} (easy question: ``What U.S. state has Atlanta as its
capital?'').
Comparable \abr{qsr} values may suggest \namedq{ambig}
and \namedq{hard} have the same difficulty.
However, \irt{}
distinguishes them: \namedq{ambig} has low
discriminability ($\gamma_j \approx 0$), resulting in a low
$p(r_{ij})$ close to 0.5 regardless of the subject skill, while
\namedq{hard} and \namedq{easy} are likely to have high
discriminability ($\gamma_j \approx 1$) and reverse difficulty
($\theta_j$) values.
\irt{} thus provides a more nuanced evaluation of
question adversarialness, capturing its appropriate challenge levels
for humans and models while accounting for its ``well-posedness''
(\S~\ref{subsec:adversarialness}).\footnote{Feasibility, another latent variable in \irt{}, also reflects poor-quality questions when a large proportion of participants answer incorrectly~\citep{rodriguez-etal-2021-evaluation}. However, our approach explicitly accounts for disagreement among highly skilled human subjects (\S~\ref{sec:ambiguity}). We leave feasibility analysis to future work.}

\newcommand{\twoplprob}[3]{\sigma_{2\text{pl}}\left(#1, #2, #3 \right)}
\newcommand{\tif}[0]{\abr{tif}}
\newcommand{\skhuman}[1]{$H_{#1}$}
\section{\advscore{}\label{section:metric}}
This section introduces \advscore{}, a metric that evaluates how
\textit{adversarial} and \textit{discriminative} a dataset is. 
We measure these two key criteria:
\textbf{(i) adversarialness}, how much more challenging a question is
for \ai{} models compared to humans while being well-posed; and
\textbf{(ii) discriminability}, how informative is the question in
effectively distinguishing between different skill levels.
%


\subsection{Quantifying Adversarialness}\label{subsec:adversarialness}
A question is adversarial if \emph{skilled} humans consistently answer
a question correctly but computers do not.
We measure this gap by fitting \irt{} parameters and then computing
the probabilities predicted by the trained \twoplirt{}
model~(\S~\ref{sec:irt}).
During margin computation, we conduct synthetic groups for both human and computer subjects with representative skill levels. 
Then, we compute the probability of each group correctly answering the question, as estimated by the \irt{} model, which accounts for question quality. A question is considered adversarial if the human representative has a higher probability of answering correctly than the computer representative.




%

\paragraph{Skilled Groups.}
We first define what constitutes a \emph{skilled} group~$g$, and
further define its \emph{representative skill} $\beta^{g}_{*}$, which
we use in subsequent equations~(\ref{eq:margin},\ref{eq:delta}).
For a set of randomly sampled subjects $S$, skilled group $S_{(k)}$ is
the subset of subjects with skill at least $k$ standard deviations
above the mean--- $\beta_i > \mu^S_\beta + k \tau^S_\beta$---where
$\mu^S_\beta$ and $\tau^S_\beta$ are the mean and standard deviation
of subject skills over the set $S$, and $k$ 
indicates the degree of expertise.
%
%
%
We define the \emph{representative skill} $\beta^{g}_{*}$ for the
chosen group $g$ as the expected skill level of the subjects within
that group:
\begin{equation}
    \beta^{g}_{*} = \mathop{\mathbb{E}}_{\beta_i \sim g}[\beta_i].
\end{equation}

\paragraph{Margin Computation.}
For question $j$ in a dataset $D$, the performance-margin
$\mu_j$ is the difference between the probabilities of \textit{skilled}
humans $H_{(0)}$ and \textit{skilled} models $M_{(0)}$ correctly answering the question,
using their respective representative skills $\beta^{H_{(0)}}$ and $\beta^{M_{(0)}}$. We set $k=0$ and
designate \emph{skilled} humans ($H_{(0)}$) and models ($M_{(0)}$) as
the skilled subsets of subjects.
These subjects have skills above the average level of their
respective subject pools:
\begin{equation}\label{eq:margin}
    \mu_j = \explain{Skilled human rep. prob.}{\twoplprob{\beta^{H_{(0)}}_{*}}{\theta_j}{\gamma_j}} - \explain{Skilled model rep. prob.}{\twoplprob{\beta^{M_{(0)}}_{*}}{\theta_j}{\gamma_j}},
\end{equation}
where $\twoplprob{\beta}{\theta}{\gamma}$ is the logistic function for
our \twoplirt{}~(Eq.~\ref{eq:2pl}, \S~\ref{sec:irt}), that uses
$\beta^{g}_{*}$ as the representative skill for subject group $g \in
\{H_{(0)}, M_{(0)}\}$, and $\theta_j$ and $\gamma_j$ are the
difficulty and discriminability parameters of the question~$j$.

\textbf{A positive value for the margin $\mu_j$ implies that the question $j$ is
\textit{adversarial}} (examples in~\ref{app:qual}), while a negative value implies the
opposite, and the magnitude indicates the extent of adversarialness.

\paragraph{Accounting for Question Ambiguity.}\label{sec:ambiguity}
While the margin ($\mu_j$) captures the core of adversarialness, it does not ensure
if the questions are genuinely well-posed; ambiguous, or poorly formulated questions
could inflate this score without being \emph{truly} adversarial.
To address this issue, we introduce a discount term (Eq.~\ref{eq:discount}) that
relies on the disagreement level among \emph{highly-skilled} (or expert) human
subjects ($H_{(1)}$) for each question:

\begin{equation}\label{eq:discount}
    \mu'_j = \frac{\mu_j}{1 + \delta_j},
\end{equation}
where $\mu'_j$ is the adjusted adversarialness score, $\mu_j$ is the original
adversarialness score, and $\delta_j$ is a measure of disagreement among highly
skilled human subjects $H_{(1)}$ for question $j$.\footnote{We use this approach for
crowdsourced human subjects. For manually identified expert human subjects,
we directly use their responses without the need for skill-based filtering.}
To keep this measure of disagreement standardized, $\delta_j$ is the mean deviation ($\text{MD}$)
of the probabilities of $H_{(1)}$ answering question $j$ correctly:
\begin{equation}\label{eq:delta}
\delta_j = \mathop{\text{MD}}_{i \sim H_{(1)}}\left[\twoplprob{\beta_i^{H_{(1)}}}{\theta_j}{\gamma_j})\right].
\end{equation}
This discount term ensures that questions with high disagreement among expert humans
(potentially ambiguous or ill-posed questions) are penalized, even
if they show large human-model performance gaps.
This approach leverages the value of human judgment for \emph{true} adversarial quality assessment.

\subsection{Measuring Discriminability}\label{subsec:disc}
The best questions distinguish between subjects' varying skill levels---they are \emph{informative}
and showcase high \emph{discriminability}. 
We measure this by leveraging Fisher information
over our \twoplirt{}'s response prediction function, also called Item Information Function~\citep[\iif{}]{lord1968statistical};
it is a function that measures an item's contribution to the measurement precision of $P(\theta)$ across the skill range ($\theta$).
With $P(\theta)$ as the \twoplirt{}'s response prediction function $\twoplprob{\beta}{\theta}{\gamma}$,
we get the item information
function ($\iif{}_j(\theta)$) that quantifies how much statistical information a question
$j$ provides about a subject's skill level $\theta$:
\begin{align}\label{eq:iif}
\iif{}_j(\theta) &= \gamma_j^2  \cdot p_j(\theta) \cdot (1 -p_j(\theta)), \text{ where} \\
p_j(\theta) &= \twoplprob{\theta}{\theta_j}{\gamma_j}.
\end{align}
%
Here, the questions with \textbf{high discrimination} (large $\gamma_j^2$) and moderate difficulty (resulting in $P(r_{ij}) \approx 0.5$) provide the most information. 

Finally, we define the total item information ($\tif{}_j$) provided by question $j$ as the area under the $\iif{}_j(\theta)$ curve, and scale it by exponential normalization to obtain a standardized, calibrated measure of discriminability $\kappa_j$ for question $j$:
\begin{align}
    \tif{}_j &= \int_{-\infty}^{\infty} \iif{}_j(\theta) d\theta, \\
    \kappa_j &= 1-\exp\left(-{\tif{}_j}\right).
\end{align}

\subsection{Combining into \advscore{}}
To recap, an ideal adversarial question should (i) have a high margin of human
and model performance gap, while being well-posed (low expert-humans disagreement), 
and (ii) be discriminative (informative of the subject's skill). Thus, first combine the 
adversarialness ($\mu'_j$) and discriminability ($\kappa_j$) to get a single
metric:
\begin{equation}
    \text{\advscore{}}_\text{j} = \frac{\mu_j}{1 + \delta_j} \cdot (1 + \kappa_j)
\end{equation}
To have human--model probability margin ($\mu_j$) as a key factor in \advscore{}, we treat $\kappa_j$ as a multiplicative bonus to $\mu_j$. This prevents questions with high discriminability ($\kappa_j$) from contributing to \advscore{} if their $\mu_j$ values are low. 

\textbf{A positive \advscore{} indicates a truly adversarial dataset, with higher values suggesting more discriminative and adversarial questions.}
We use \advscore{} to evaluate existing datasets (\S~\ref{sec:comparison}) and to reward authors in our \advqa{} dataset creation process (\S~\ref{sec:adversarial-competitions}). We define the \advscore{} of a dataset $D$ as the average \advscore{} of its questions. An effective adversarial dataset should contain numerous questions with high \advscore{}.

\section{Adversarial Benchmark Evaluation} 
\label{sec:comparison}

\definecolor{DarkOrange}{RGB}{255,140,0}
\definecolor{OliveGreen}{RGB}{26,165,10}
\definecolor{TealBlue}{RGB}{0,100,153}
\definecolor{VioletRed}{RGB}{204,0,50}
\definecolor{Plum}{RGB}{153,0,153}
\begin{figure*}[!h]
    \centering
    \includegraphics[width=1.0\linewidth]{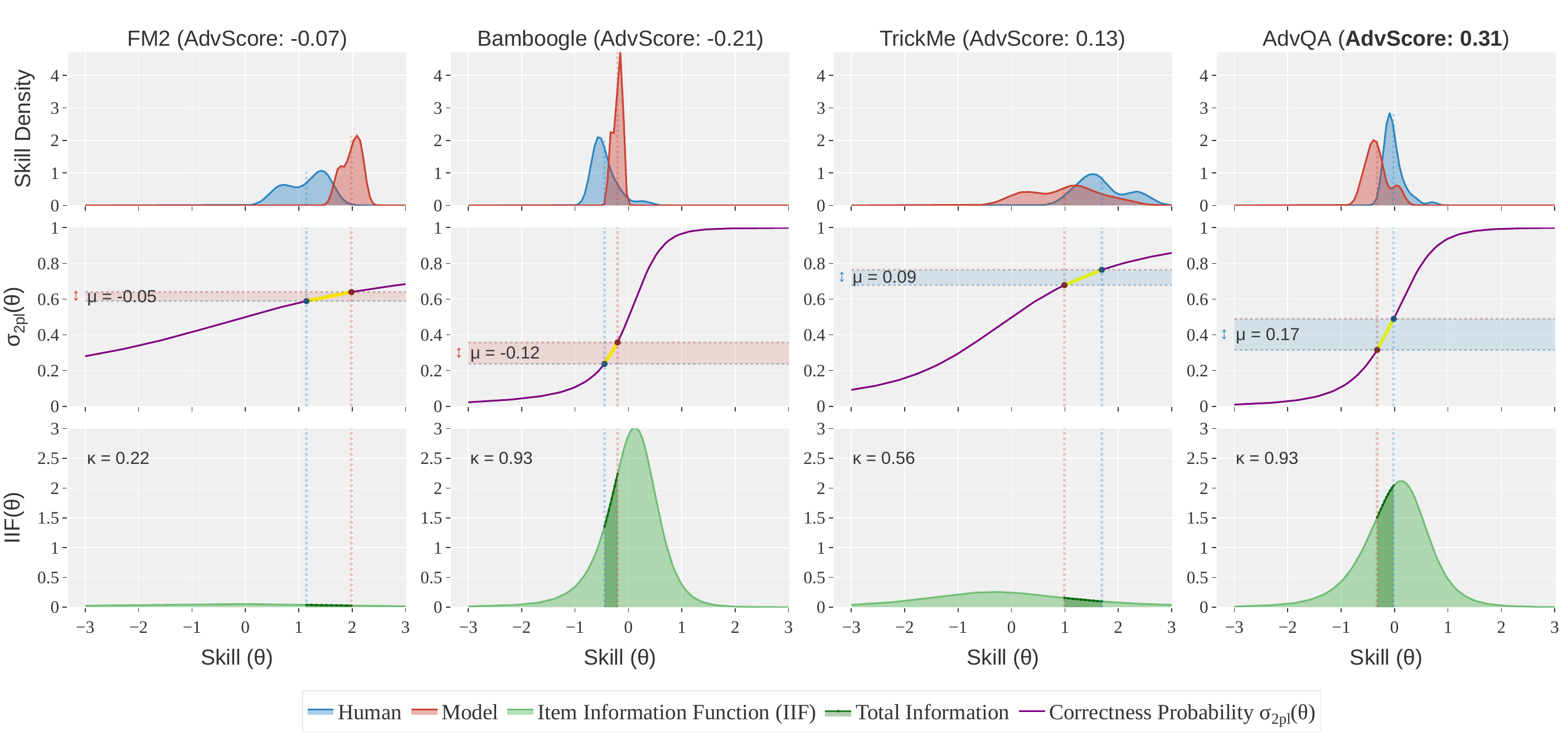}
    \caption{\textbf{Visualization of key \advscore{} components across datasets.} For each dataset, we plot:
    (1) Skill density of \textcolor{TealBlue}{skilled humans~($H_{(0)}$)} and \textcolor{VioletRed}{skilled models~($M_{(0)}$)},
    (2) \textcolor{Plum}{response correctness probability, $\sigma_{\text{2pl}}(\theta)$} (Eq.~\ref{eq:2pl}, \S~\ref{sec:irt}) averaged over dataset examples, and
    (3) \textcolor{OliveGreen}{Item information function ($\iif{}(\theta)$}(Eq.~\ref{eq:iif}, \S~\ref{subsec:disc}).
    Vertical dashed lines show representative (average) skill levels for humans and models.
    The gap between human and model probabilities (shaded region between the horizontal lines)
    indicates adversarialness ($\mu_D$). \iif{} peaks show where questions are most informative, with 
    area under curve signaling total informativeness (discriminability, $\kappa_D$). 
    \textbf{Key insights:} \abr{bamboogle} has high informativeness but favors models (negative $\mu_D$). 
    \abr{TrickMe} separates humans and models but has lower discriminability (positive $\mu_D$). 
    \advqa{} is the best of all, effectively discriminating between humans and models 
    while maintaining high informativeness throughout, resulting in the highest \advscore{} of 0.31.}
    \label{fig:skill_density}
\end{figure*}
We compare adversarial benchmarks across different domains using \advscore{}.
Our evaluation includes \advqa{}, a new \qa{} dataset developed through a
human-in-the-loop~(\hitl{}) process to align adversarial data with human capabilities. 
This section, analyzes \advscore{} as a metric, while \S~\ref{creationpipeline} details
the creation of \advqa{}, and \S~\ref{sec:analysis} examines what makes \advqa{}
questions adversarial.

\paragraph{Adversarial datasets with human responses.} For \advqa{}, we gathered human
responses through a live, in-person \qa{} competition involving 8 human teams, as well
as through online crowdsourcing with 165 participants. In total, we collected 1,839
human responses from 172 individuals.
To compare the adversarialness of these datasets using \advscore{}, which relies on both
human and model response data, we are limited to comparing \advscore{} with datasets with
human annotations. Thus, we select \abr{trickme}~\citep{wallace2019trick} and
\abr{fm2}~\citep{eisenschlos-etal-2021-fool}. While \abr{trickme} challenges models with 
\qa{} pairs, \abr{fm2} uses entailment pairs for fact-checking.\footnote{We use human
responses from \citet{si2023large}} Additionally, we included \abr{bamboogle}~\citep{press2022measuring},
which consists of general knowledge questions designed to be adversarial, similar to \advqa{}.
As \abr{bamboogle} lacked human responses, we gathered 10,391 responses from 165 crowdworkers. 

We also collected model responses for each dataset from ten models, including Dense Passage Retrieval
(\abr{dpr})~\citep{karpukhin-etal-2020-dense}, \abr{GPT-3-Instruct}~\citep{ouyang2022training},
\abr{gpt-3.5-turbo}~\citep{openai2023chatgpt}, \abr{mistral-v0.1-instruct}~\citep{jiang2023mistral},
\abr{gpt-4}~\citep{achiam2023gpt}, \abr{llama-2-chat} models in sizes of 7b and 70b, and
\abr{llama-3-instruct} models in sizes of 8b and 70b~\citep{touvron2023llama}. After collecting
human and model responses, we apply \twoplirt{} to extract the learned subject and item
parameters and compute \advscore{}.

\paragraph{Comparison of adversarial benchmarks.}
We compute $\advscore{}_D$ and its components ($\mu_D$, $\kappa_D$, and $\delta_D$) for
each dataset, presenting results in Table~\ref{tab:advscore_specifics}.
Figure~\ref{fig:skill_density} walks through the computation of \advscore{} by illustrating
(i) the skill density of \emph{skilled} humans $H_{(0)}$ (blue) and models $M_{(0)}$ (red),
(ii) the response correctness probability ($\sigma_{\text{2pl}}$, purple), and
(iii) the \emph{item information function}, \iif{} (green, E.q.~\ref{eq:iif}), over skill $\theta$.

Both \advqa{} and \abr{trickme} show a clear separation between human and model skill levels (first row),
resulting in positive, high margins ($\mu$) of 0.17 and 0.13, correspondingly (yellow in second row). However, \advqa{} has a
higher overlap of \iif{} with regions where human skill exceeds model skill (dark green area in third row), compared to \abr{TrickMe},
which has a flatter and less informative \iif{}. These lead to lower $\kappa_D$ (0.56 vs 0.93),
suggesting that \abr{trickme} questions are less discriminative (less useful in assessing subject skills). 

In contrast, \abr{bamboogle} has an informative \iif{}, but the skill of the model tends to exceed humans, resulting in a negative $\mu_D$ (Table~\ref{tab:advscore_specifics}).
This suggests that \abr{bamboogle} questions are inversely adversarial, containing questions where models outperform humans, and therefore fail to serve as an effective adversarial benchmark.
Similarly, \abr{fm2} has a negative $\mu_D$ and low $\kappa_D$,
indicating that the dataset is neither adversarial nor discriminative.
%
Our analysis establishes \advqa{} questions as most adversarial, as indicated by its highest $\advscore{}_D$ of 0.31; thus demonstrating that the unique components of \advscore{} effectively support the evaluation of adversarial benchmarks.
\begin{table}[!t]
    \centering
    \footnotesize
    \setlength{\tabcolsep}{3pt} 
    \renewcommand{\arraystretch}{1.2}
    \begin{tabular}{lrrrr} 
    \toprule
         \textbf{Datasets ($D$)} & \textbf{$\mu_D$} & \textbf{$\kappa_D$} & \textbf{$\delta_D$} & \textbf{$\advscore_D$} \\
    \midrule
        \textbf{\advqa{}}   & \textbf{0.17}  & \textbf{0.93}  & 0.08 & \textbf{0.31} \\
        \abr{fm2}           & -0.05          & 0.22          & 0.01          & -0.07         \\
        \abr{bamboogle}     & -0.12          & 0.93           & \textbf{0.11}          & -0.21         \\
        \abr{trickme}       & 0.09           & 0.56           & 0.03          & 0.13          \\
    \bottomrule
    \end{tabular}
    \caption{\advqa{} had the highest $\advscore_D$, along with the highest $\mu_D$ and $\kappa_D$, indicating that its questions were the most adversarial and best at discriminating \subj{}'s skill across the four datasets. While \abr{bamboogle} has the same $\kappa_D$ value, the negative $\mu_D$ indicates the reverse adversarialness, suggesting it was distinctively easier for \textit{models} than humans.}
    \label{tab:advscore_specifics}
\end{table}

\begin{figure}[!t] 
    \centering
    \includegraphics[width=\linewidth]{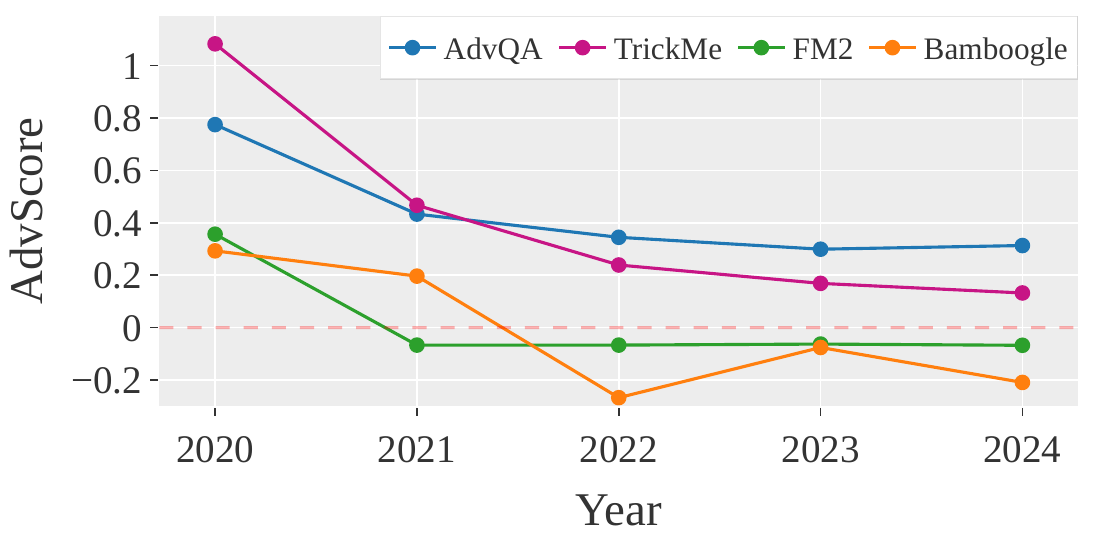}
    \caption{We report \advscore{} for each dataset over the years,
    confirming that \advqa{} holds the highest \advscore{} with the
    smallest decline over the last five years, proving its adversarial robustness.}
    \label{fig:cumulative_advscore}
\end{figure}

\paragraph{Chronological evaluation of adversarialness}
Adversarial datasets inevitably become obsolete as models improve, either by training on
these datasets or overcoming previously identified vulnerabilities. Using \advscore{},
we assess model improvements over the last five years by identifying which datasets have
become less adversarial, incorporating new models into the \advscore{} computation.\footnote{Models
introduced by year: DPR in 2020, GPT-3-Instruct in 2021, GPT-3.5-TURBO in 2022,
Mistral-0.1-instruct, GPT-4, Llama-2-7b-chat, and Llama-2-70b-chat in 2023, and
Llama-2-7b-chat, Llama-2-70b-chat, Llama-3-8b-instruct, Llama-3-70b-instruct, and
rag-command-r-plus in 2024.} 
Figure~\ref{fig:cumulative_advscore} shows the \advscore{} for each dataset over the years,
confirming that \advqa{} holds the highest \advscore{} (2024) with the smallest
decline over the last five years.
In contrast, \abr{trickme}, which was initially the most
highly adversarial (2020), saw a sharp decline over the following four years,
indicating that the models improved on the tasks that they previously struggled with.
\abr{bamboogle} and \abr{fm2} are no longer adversarial, showing negative \advscore{} values since 2022. \abr{bamboogle}'s reliance on a 2-hop tactic and simple questions (e.g., {\small\textit{``What is the capital of the second largest state in the US by area''}}) likely explains its decline since 2021. \abr{fm2}'s drop suggests \abr{llm}s have improved at fact-checking or benefitted from similar questions in training.
Although pinpointing the exact factors behind model improvement may be challenging, it is crucial to determine whether these models
have become more resilient or remain vulnerable as new models emerge. \advscore{} facilitates this by quantifying how much a dataset has lost its
adversarialness, offering a concrete measure of how well the model withstands adversarial challenges over time.

\paragraph{Qualitative Examples with \advscore{}}\label{app:qual}
We examine the human-model margin probability ($\mu_j$) and each subject's answers to the example question for each dataset. 
In Table~\ref{tab:adv_example}, \advqa{} and \abr{trickme} questions show a positive $\mu_j$ value, indicating adversarial, correspondent to the human's correct answer to (``Putin'') and GPT4's wrong answer (``Russia''). 
On the other hand, \abr{bamboogle} and \abr{fm2}'s negative adversarialness value suggests that the question is easier for models compared to humans, as reflected in the higher correctness from models versus humans.

\paragraph{Comparision of \advscore{} and \abr{QSR}} Moreover, we conducted a comparative analysis of model and human success rates (\abr{QSR}) and \advscore{}s (\S~\ref{subsec:qsr}). While \abr{QSR} may suggest that humans outperform models, the questions can consistently yield negative \advscore{}s, due to their low or negative $\mu$ (margin) or high $\delta$ (ambiguity). Examples and analyses in  Appendix~\ref{qsr_advscore_comparison}). This highlights that \abr{QSR} alone is insufficient to determine question adversarialness, whereas each parameter in \advscore{} offers a more reliable measure.

\begin{table*}[!t]
    \centering
    \scriptsize
    \setlength{\tabcolsep}{7pt} 
    \renewcommand{\arraystretch}{1.0} 
    \begin{tabular}{p{15mm}p{50mm}p{12mm}p{14mm}p{20mm}p{15mm}}
    \toprule
        \textbf{Dataset} & \textbf{Question} & \textbf{Answer}  & \textbf{Margin ($\mu_j$)} & \textbf{Human Response}  & \textbf{\gpt{}-4}  \\ 
    \midrule
        \advqa{} & Who is the president of the country represented by the second letter in the acronym BRICS [...]  & Vladimir Putin  & 0.19  &  Putin & Russia  \\
    \midrule 
        \abr{fm2} & Aram Khachaturian had Russian roots. & False  &   -0.01   & ``False'' & True  \\
    \midrule
        \abr{trickme} &In a novel by this author, a detective wraps his arm to survive a dog attack [...]  & Durrenmatt  & 0.12 &   ``Durrenmatt'' & Franz Kafka\\
    \midrule
        \abr{bamboogle} & Who directed the highest grossing film?  & James Cameroon & -0.02 & ``No idea'' &  James Cameron \\ 
    \bottomrule
    \end{tabular}
    
    \caption{\advqa{} demonstrates the most balanced properties of challenging the model and distinguishing between skills, as indicated by a positive $\mu_j$ value, which aligns with humans outperforming the models. }
    \label{tab:adv_example}
\end{table*}

\section{\advqa{} creation pipeline} \label{creationpipeline}
In the previous sections, we showed that \advqa{} is more adversarial and discriminative than other datasets, suggesting its creation process contributed to these qualities. Here, we discuss the \advqa{} collection process as a case study to guide future high-quality adversarial datasets.

\subsection{Collecting questions and answer pairs through adversarial competitions}
\label{sec:adversarial-competitions}
To obtain human-written question-answer pairs, we hold two adversarial model--human \qa{} competitions. 
First, in the writing competition, we collect 399 adversarial questions through the interface (\S \ref{interface}), which are then edited and filtered by an expert editor.
Second, in the answering competition, we invited eight expert human groups (composed of three to four trivia experts) to run eight human vs. model \qa{} tournaments to obtain 780 human responses. Each tournament initially consisted of 30 questions, which are then filtered based on experts' comments (E.g., {\small{\emph{``This question is ill-posed''}}}). 
After this filtering process, \advqa{} results in 182 questions.\footnote{
Larger than other \irt{}-analysed test sets (e.g., 139 for \abr{RTE}, 20 for \abr{CommitmentBank}, 50 for \abr{COPA})~\citep{vania-etal-2021-comparing}. Also, additional 1,839 human responses collected from 172 individuals (165 crowdsource workers). Dataset value includes both questions and response volume.}
After the competitions, we incentivize the writers with the highest \advscore{} and players with the highest skill.\footnote{\advscore{} is not computed \textit{during} the dataset construction. It is a post-hoc evaluation metric.}

\subsection{Skilled writers use adversarial interface}\label{interface}
We provide an adversarial writing interface as a human-AI collaborative tool for the adversarial writing competition, motivated by~\citet{you-lowd-2022-towards}'s finding that human-AI collaboration strengthens adversarial attacks. We supply the writers with real-time model interpretations, inspired by~\citet{wallace2019trick}; they could continuously counteract the model response and make edits.

\paragraph{Eliciting incorrect model predictions} The center of the interface (Figure  \ref{fig:interface} in Appendix \ref{interface_screenshot}) provides the Wikipedia page for the target answer, which they use to write the question. While the author is writing, the retrieval widget and \qa{} models widgets are updated ~\citep{eisenschlos-etal-2021-fool}.
Motivated by \citet{feng2018pathologies}, we embed the input perturbation inside the question writing widget to highlight which words trigger the model
predictions. For example, changing ``company'' to a different token would be most likely to change the prediction except the answer ``Apple.''
\paragraph{Retrieval systems} \label{retrieval}
Users receive real-time feedback on \qa{} systems' performance on their questions via the interface's fine-tuned retrieval and reader model components (the retrieval system outputs: contexts that elicit \qa{} system predictions). If the target answer appears at the top of the retrieval widget, which means the author failed to 
fool the retriever and the reader, authors can rephrase questions to avoid retrieving information that makes \qa{} systems answer correctly. 
We use lightweight sparse and neural retrieval models for writer feedback: a \abr{tf-idf} baseline and \abr{dpr}. To ensure that \abr{dpr} predictions are diverse and up-to-date, we create a database that indexes each sentence in a set of Wikipedia pages (see Appendix~\ref{retrieval}). 
We then use the \roberta-based \farmreader{}, which is fine-tuned on \squad ~\citep{rajpurkar-etal-2016-squad}, to read and sort the retrieved sentences from the two retrieval models by their relevance. 
\paragraph{\textbf{\abr{lm}-based \qa{} systems}}
We enrich the model guidance using extractive and generative model answer predictions. For extractive \qa{}, we use \distilbert{} (fine-tuned on \squad), since its promptness and lightness facilitate rapid human-AI interaction. We
also use \abr{T5}\footnote{The writing competition was held in Spring 2023, when \distilbert \,and T5 were considered comparatively strong.}~\citep{raffel2020exploring} to answer the questions in a closed-book setting.

\section{Discussion and Analysis on \advqa{}} \label{sec:analysis} In this section, we show how \advscore{} can help identify factors that encourage high-quality adversarial datasets. Effective strategies in \advqa{} may guide the creation of more adversarial questions, and we analyze how the dataset's realistic aspect can help incorporate human variability during model evaluation.

\paragraph{Ensuring high-quality adversarial questions} The questions should be adversarial for reasons that identify model weaknesses, such as the inability to compose clues or exclude redundant clues~\citep{min2020ambigqa, min2022crepe} not because of trivial errors (e.g., grammar mistakes). If the question meets this criteria, we consider it high-quality. We base our criteria on the taxonomy of adversarial categories in~\citet{wallace2019trick}. To understand what yielded \advqa{}'s \textit{high-quality} adversarial questions, manually annotate the adversarial tactics and topics for \advqa{} questions (Appendix~\ref{app:advtype}). 

\begin{figure}[!t] 
    \centering
    \includegraphics[width=\linewidth]{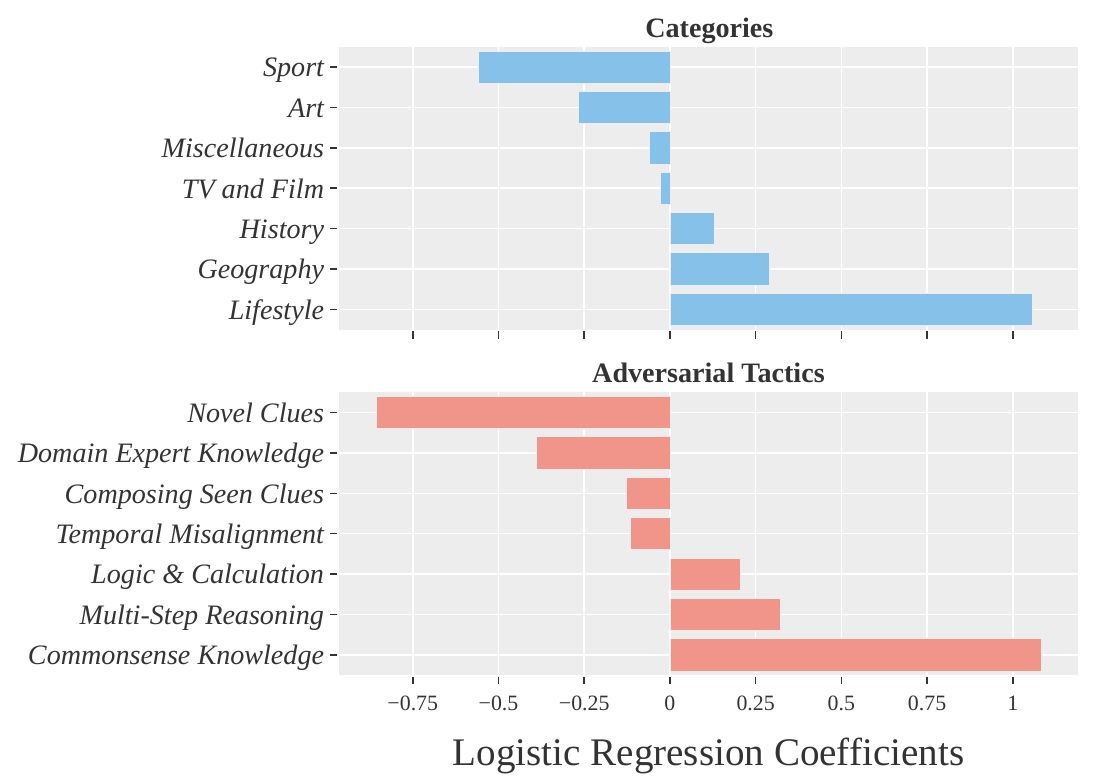}
    \caption{The overall distribution of LR coefficients suggests that \textit{lifestyle} and \textit{commonsense knowledge} contribute more to adversarialness than other features. This implies that models still struggle with commonsense knowledge, highlighting an area where they remain vulnerable compared to human understanding.}
    \label{fig:feature_weight}
\end{figure}

With the identified question characteristics, we run a logistic regression model to learn how much each adversarial tactic or topic contributed to \advscore{}.\footnote{Focusing on assessing adversarialness through \irt{}, we provide only a basic analysis using pre-assigned features. Applying advanced \irt{} models is encouraged for a richer analysis of adversarial factors~\citep{gor2024great}.} Since all questions in \advqa{} yielded a positive \advscore{}, the coefficients in Figure~\ref{fig:feature_weight} reflect how much specific features contributed to adversarialness, highlighting areas where models need improvement. For instance, the tactic involving \textit{commonsense knowledge} on the topic of \textit{lifestyle} exposed a model weakness (e.g., \textit{\small{“Take away four from a group including Barnard and Smith, and you get what play?”}}), which had a notably high \advscore{} of 0.27.\footnote{The low number of \textit{TV \& Film} questions, likely tied to recent news, confirms that \advqa{} focuses on probing model capabilities rather than time-sensitive knowledge (Appendix~\ref{app:advtype}).} 
\setlength{\textfloatsep}{10pt plus 1.0pt minus 2.0pt}
\paragraph{Leveraging human feedback for \textit{realisticness}} 
Realism is crucial for an adversarial dataset as it creates challenges that closely resemble real-world scenarios, effectively testing model robustness against plausible but diverse situations. This approach enhances the reliability of performance evaluation as it reflects high variance in collective human ability. For example, not only should the questions be adversarial, but they should mimic diverse reasoning and problem-solving strategies of different people.
Our preliminary results revealed that crowdworkers often produced ambiguous or poorly-formed questions.\footnote{E.g., \textit{\small{"Who led the final siege of Constantinople?"}} carries ambiguity depending on historical framing (\textit{\small{Mehmed II for the 1453 siege}} or \textit{\small{other leaders in prior sieges}}).} Although \advscore{} could identify these issues, many examples were ineffective for assessing model performance. We thus recruit expert trivia writers and guide them in writing adversarial questions. Then, other trivia editors scrutinize the human-authored questions' poor quality (see Appendix~\ref{app:questiontype}). Finally, our human vs. model competition provides an additional quality check, as human \subj{}s flag potential issues while answering questions. If the \subj{} or the editor considers a question unnatural or ambiguous, we exclude it from our final dataset (Appendix~\ref{triviadetails}). 

We emphasize that human responses are especially useful in adversarial evaluation contexts, as they ensure that adversarial examples are genuinely challenging and realistic. Moreover, these responses are provided by each individual's intuition, creativity, and understanding. Thus, capturing variability is crucial to evaluate the benchmarks that are meant to assess evolving models aiming for human alignment. Such aspects are what  traditional model-generated adversarial attacks cannot replicate. Ultimately, incorporating human responses adds depth and reliability to adversarial benchmarks, making them essential in evaluating models' true progress toward human-level understanding and their performance.





\section{Related Work}
Adversarial samples expose and evaluate model capabilities~\citep{melis2017deep,biggio2013evasion}. 
Recently, the Natural Language Processing (\abr{NLP}) community has questioned whether models trained on
benchmarks learn to solve tasks in robust and generalizable
ways~\citep{ribeiro-etal-2020-beyond,bartolo2021improving, Nie2018AnalyzingCO,
  gururangan-etal-2018-annotation,kaushik-etal-2021-efficacy}. Thus, evaluation of adversarial samples has been active in areas of reading comprehension~\citep{jia2017adversarial} and neural translation tasks~\citep{belinkov2018synthetic, wallace-etal-2019-universal}.
\citet{tedeschi-etal-2023-whats}
postulates that the abilities of many ``superhuman'' models may be overestimated due to 
poorly annotated datasets and biases embedded in the evaluation process (e.g., fixed test sets). 

An alternative is to provide more challenging benchmarks that require a stronger form of generalization and diversity~\citep{10.1007/978-3-030-36718-3_20, bowman2023eight, yuan2023tasklama}; \abr{hitl} adversarial generation framework enables humans create examples while interacting with the model~\citep{Ma2021DynaboardAE}.
For \qa{} tasks, it is crucial to validate the model's ability to correctly answer easy and natural questions that are likely to be expressed by humans. For \abr{hitl} adversarial generation for \qa{}, \citet{bartolo2021improving} and \citet{kiela-etal-2021-dynabench} uses a synthetic
generation method to amplify small set of human-authored adversaries. \citet{adversarialvqa}
introduces a benchmark in which the humans interact with a visual \qa{}
model, and write an adversarial
question for each of a set of images. \citet{wallace2019trick} and \citet{eisenschlos-etal-2021-fool} both use \abr{hitl} incentive mechanisms to create adversarial questions. For evaluation of these adversarial datasets, 
\citet{lalor2019learning} introduces an \abr{irt}-based ranking
method to remedy the issue that current evaluation treats each model independently rather than considering relative differences. \citet{rodriguez-etal-2021-evaluation} also redesigns the
leaderboard framework with a Bayesian approach where latent subject
skill and item difficulty predict correct responses.  
Our \advscore{} can systematically probe models to understand their
capabilities, and provide a measure to understand which 
also contribute in \abr{hitl} adversarial dataset framework to help to create the next generation of data.

\section{Conclusion}\label{conclusion}
Adversarial datasets offer practical benefits for evaluating models to improve robustness and performance. Grounded in human feedback, \advscore{} ensures that evaluations of adversarial benchmarks align with human capabilities by post-hoc assessment of adversarial robustness and model improvements. 
Thus, applying \advscore{} in real-time benchmark construction can aid in evaluating the robustness of the models, and integrating \advscore{} into model training can improve their adaptability to real-world applications. 
\newpage
\section{Limitations and Future Works}\label{limitation}
One limitation of \advscore{} is its reliance on expert-level human annotations that makes it challenging to implement. However, human feedback ensures that adversarial questions are not only technically challenging but also meaningful and reflective of real-world scenarios. To mitigate this, semi-supervised or active learning approaches could be explored to minimize manual annotations, where models assist in identifying adversarial examples based on human feedback.

Another limitation is that \advscore{} does not account for model confidence, which may overlook reliability aspects. We recommend incorporating a calibration assessment to determine if predicted probabilities align with accuracy, encouraging more reliable adversarial benchmarks and thereby preventing overconfident models.

Furthermore, as the core of \advscore{} aims to assess how well models match human ability in real-life tasks, it is valuable to evaluate adversarial datasets in real-world applications, such as machine translation and chatbot evaluation across different modalities. We encourage using \advscore{} to develop adversarial datasets across diverse NLP tasks and contribute to robust system developments.


\section{Ethical Considerations} \label{Ethics}
We address ethical considerations for dataset papers, given that our work contains a new dataset \advqa{} and collecting human responses in our user study. We reply to the relevant questions posed in the {\texttt{\abr{acl} 2022 Ethics \abr{faq}}}\footnote{https://www.acm.org/code-of-ethics}. 

When collecting human responses and questions, our study was pre-monitored by an official \abr{irb} review board to protect the participants' privacy rights. Moreover, the identity characteristics of the participants were self-identified by the workers by answering the survey questions. 

Before distributing the survey, we collected consent forms for the workers to agree that their answers would be used for academic purposes. 
The trivia experts were awarded a total $\$1100$ worth of online gift cards after the competitions. The prizes were awarded to the first, second, and third winners, depending on each \team{}'s \advscore{}.
The crowdworkers were compensated over $10$ \abr{usd} an hour (a rate higher than the \abr{us} national minimum wage of $7.50$ \abr{usd}
).

\section{Acknowledgements}
We thank all the CLIP members who reviewed the idea of improving adversarial benchmark evaluation. We also thank the players who participated in the tournament: Munir Siddiqui, Aaron Lichtig, J.R. Parsons, Ethan Medwetsky, Matt Weiner, and Alex Schmidt. Their valuable contributions greatly impacted the progress of this work. This project was awarded the MetaAI Dynabench Grant ``A Leaderboard and Competition for Human–computer Adversarial Question Answering''. Additionally, this research was partially supported by an NSF GRFP grant. Sung and Boyd-Graber are supported by NSF Grant IIS2403436. Opinions, findings, conclusions, or recommendations expressed here are those of the authors and do not necessarily reflect the views of the sponsors.

\bibliography{bib/yy}
\bibliographystyle{style/acl_natbib.bst}
\newpage
\newpage
\appendix
\section{Details on Dataset Creation}
\label{appendix:dataset}
\subsection{Recruitment for Dynamic \abr{qa} Generation}\label{triviadetails}
When tasking human authors with adversarial writing of questions, \citet{wallace2019trick} emphasizes the importance of ``who'' the authors should be: \textit{talented and eager} question writers with \textit{specific goals}; they should aim to generate questions that stump computers but seem normal enough for humans to answer. To make this work, they recruit members of the quizbowl community, who have deep trivia knowledge and craft question for quizbowl tournaments~\citep{jennings2007brainiac}. However, their challenge was to convey what is "normal" to authors and stimulate examples that can elucidate the weaknesses of \abr{qa} models.

\subsection{Merging Trivia Question Generation and Dynamic Adversarial Generation Process} \label{merge}
Many QA datasets are now too easy for modern models as models have become more powerful~\citep{rogers2023qa}. However, even these easy QA datasets have serious data flaws~\citep{min2020ambigqa, yu-etal-2023-crepe}, which suggests that creating question-answer pairs is a very challenging task. This is also a norm for questions written for human players, where more than 100,000 questions are produced annually. To create effective and challenging enough questions, the professional experts (e.g., writing staff) take a rigorous editing pass on the questions to decide whether they are adequate enough to guarantee players a fair game~\citep{lelkes2021quiz, pollard2006student}. They follow strict guidelines to be selected to be used in the quiz matches. We propose to merge the above pipelines to help improve data creation for robust QA models by adding an editing step to ensure that grammatical errors and nonfactual questions (following the norms of Trivia questions) do not exist in the pool. In Table \ref{badquestion}, we list the problematic question types that we ask the editors or \subj{}s to flag. 
\begin{table*}[!h]
    \centering
    \scriptsize
    \setlength{\tabcolsep}{7.5pt} 
    \renewcommand{\arraystretch}{1.8}
    \begin{tabular}{p{30mm}p{30mm}p{50mm}}
    \Xhline{1pt}
         \textbf{Question Type} & \textbf{Description} & \textbf{Examples} \\ 
    \Xhline{1pt}
        {Lacks Factuality}  &  Requires information is factual
 & {``Trump, the first woman president of the United States, is charged against federal laws'' is non factual as the gender of Trump is male} \\
        {Lacks Specificity \newline (False Presupposition)}  &  Requires more information to be answered with clarity  & {'What is the color of Flamingo’s feathers?' is ambiguous as Pink and White could be two possible answers depending on when they are born
 } \\
        {Subjectivity}  &  Contains clues that are highly subjective & {``What’s the name of Christopher Columbus’s most famous ship?'' Possible answers could be either Santa Maria, La Nina, Santa Clara. Also, as ``Most famous'' can mean many different things, the revised question could be ``Which of Columbus’s ships was stripped of its timbers to build a fort called La Navidad in northern Haiti?''
} \\
        {Ambiguity \& \newline Multiple acceptable answers
}  &  Can be answered with multiple answers & 
        {Nikolas Alexandrovitch Romanov, Nikolas II, Nikolai II Alexandrovich Romanov: all of these are acceptable as answers.} \\
        \Xhline{1pt}
    \end{tabular}
    \caption{We list the problematic question types that we ask to annotate. The four types are illustrated with descriptions and examples to help them better understand each question, and help determine whether each question has good quality.}
    \label{badquestion}
\end{table*}

\subsection{Details on errors in using raw scores in question answering competition} \label{accuracynogood}
We infer that
the human accuracy does not necessarily translate to answering ability or question difficulty measurement, which
obscures the measuring the the question’s adversarial-ness. While the most skillful human team answered all three questions correctly, the estimated probability of
the human teams answering the question correctly when compared to their ability was low (50\%). 
\begin{table*}[!t]
    \centering
    \small
    \setlength{\tabcolsep}{8pt} 
    \renewcommand{\arraystretch}{1.4}
    \begin{tabular}{p{75mm}p{20mm}p{10mm}p{15mm}}
    \Xhline{1pt}
         Question & Gold Answer & Human Answer &  Probability $\sigma(\beta_i-\theta_j)$\\ 
    \Xhline{1pt}
        What phrase is common to the title of novel featuring a fictional Nat King Cole recording, a Gene Autry film and song, and an I-95 attraction between the Carolinas?  &  South of the Border & Correct &  0.57 \\
        In which novel, written by an author who was originally a botanist and born in Cuba, features a fictitious conversation between a merchant who travelled a road that was known by a smooth natural material and an emperor who loved to write Chinese poetry, both of which are actual people in history? & Invisible Cities & Correct &  0.55\\
        What is the name of the first mosque in the world that was built by Prophet Muhammed (s.a.w) during his hijrah from Mecca to Medina?  & Quba Masjid & Correct & 0.56  \\
        \Xhline{1pt}
    \end{tabular}
    \caption{While the most skillful human team answered all three questions correctly, the estimated probability of the human teams answering the question correctly when compared to their ability was low (50$\%$).}
    \label{table:advex}
\end{table*}

\subsection{Qualitative Examples of each dataset with \advscore{}}\label{app:qual}
We examine the adversarial properties of each question ($\mu_j$ and $\kappa_j$) with qualitative examples and each subject's example responses from four datasets (Table~\ref{tab:adv_example}).

\subsection{Comparison Analysis of \advscore{} and \abr{QSR}}\label{qsr_advscore_comparison}
We show that \abr{QSR} alone is insufficient to determine question adversarialness, obscuring the real challenge,  whereas each parameter in \advscore{} offers a more nuanced measurement.

For questions like \textit{What was the founding date of the university in which Plutonium was discovered?} and \textit{Who is the father of the father of observational astronomy?}, humans significantly outperformed models, but their negative  \advscore{}s (\(-0.365\) and \(-0.340\)) indicate that these questions remain non-adversarial. This demonstrates that QSR alone is insufficient to identify question adversarialness. \advscore{}, by incorporating both margin and discriminative power, provides a more nuanced and reliable measure, and reflects the adversarial nature of questions.

\begin{table*}[h!]
\centering
\scriptsize
\setlength{\tabcolsep}{6pt} 
\renewcommand{\arraystretch}{1.5}
\begin{tabular}{@{}p{5cm}p{2cm}ccccccc@{}}
\toprule
\multicolumn{8}{c}{\textbf{AdvQA Dataset}} \\ \midrule
\textbf{Question} & \textbf{Answer} & \textbf{Human QSR} & \textbf{Model QSR} & \textbf{$\mu_j$} & \textbf{$\delta_j$} & \textbf{\(\kappa_j\)} & \textbf{\(\advscore_j\)} \\ \midrule
Name the color of the sky in Aivazovsky's ``The Ninth Wave'' & Orange & 0.667 & 0.083 & 0.583 & 0.106 & 0.963 & 0.188 \\
The title of this book shares a word with the title of a song of which the author, who acted in the 2002 film, 8 Mile, addressed to his daughter and niece & To Kill a Mockingbird & 0.333 & 0.000 & 0.323 & 0.102 & 0.983 & 0.179 \\
What country shares a language with its more populous northern neighbor but in its written form omits a letter that looks like a Greek beta, writing the sound instead by doubling another letter? That character appears in that language's words for foot, big, outside, and street & Switzerland & 0.333 & 0.000 & 0.333 & 0.051 & 0.626 & 0.081 \\
A German admiral sailing for Russia named what islands for an English captain and not for the librettist of the HMS Pinafore nor for the announcer of Jeopardy! & Gilbert Islands & 0.333 & 0.100 & 0.233 & 0.034 & 0.504 & 0.051 \\ \midrule

\multicolumn{8}{c}{\textbf{Bamboogle Dataset}} \\ \midrule
\textbf{Question} & \textbf{Answer} & \textbf{Human QSR} & \textbf{Model QSR} & \textbf{$\mu_j$} & \textbf{$\delta_j$} & \textbf{\(\kappa_j\)} & \textbf{\(\advscore_j\)} \\ \midrule
What was the founding date of the university in which Plutonium was discovered? & March 23, 1868 & 0.452 & 0.167 & 0.285 & 0.127 & 0.972 & -0.365 \\
Who was the father of the father of psychoanalysis? & Jacob Freud & 0.528 & 0.500 & 0.028 & 0.149 & 0.982 & -0.354 \\
When did the person who gave the Checkers speech die? & April 22, 1994 & 0.200 & 0.167 & 0.033 & 0.156 & 0.985 & -0.350 \\
Who is the father of the father of observational astronomy? & Vincenzo Galilei & 0.324 & 0.167 & 0.157 & 0.121 & 0.964 & -0.340 \\
What is the third letter of the top-level domain of the military? & l (lower case L) & 0.516 & 0.333 & 0.183 & 0.152 & 0.983 & -0.338 \\ \bottomrule

\end{tabular}
\caption{A substantial gap in \abr{QSR} may suggest human superiority over models, indicating an adversarial question. However, it can still yield negative \advscore{}s due to low or negative $\mu$ or relatively high $\delta$. In both \advqa{} and Bamboogle, even when human \abr{QSR} surpasses model \abr{QSR}, this is not always reflected in \advscore{}, given the distinct criteria of each parameter. For instance, the first question in \advqa{}, \textit{Name the color of the sky in Aivazovsky's ``The Ninth Wave''} exhibits a significant \abr{QSR} gap between humans (0.667) and models (0.083), yet its positive \(\advscore_j = 0.188\) remains low, due to high $\delta$ (indicating question ambiguity) compared to other examples. The question implies a single color, but the ``The Ninth Wave'' painting contains multiple hues. It also lacks specificity about which part of the sky is being referenced.}
\label{tab:qsr_table}
\end{table*}

In \advqa{}, \advscore{} highlights contrasts that \abr{QSR} may fail to capture. For instance, the question \textit{Name the color of the sky in Aivazovsky's ``The Ninth Wave''} exhibits a significant QSR gap between humans (0.667) and models (0.083), yet its positive \(\advscore_j = 0.188\) remains low, due to high $\delta$ (indicating) compared to other examples. The question implies a single color, but the `The Ninth Wave'' painting contains multiple hues. It also lacks specificity about which part of the sky is being referenced.

Other examples in Table~\ref{tab:qsr_table} show a similar trend of having a high \abr{qsr} gap, suggesting that humans significantly exceed model performance, but this is contradicted by the corresponding \advscore{}. For example, the question \textit{What country shares a language with its more populous northern neighbor but in its written form omits a letter that looks like a Greek beta, writing the sound instead by doubling another letter?} shows low discriminability (\(\kappa_j = 0.626\)) and a low \(\advscore_j = 0.081\). 
The question \textit{A German admiral sailing for Russia named what islands for an English captain and not for the librettist of the HMS Pinafore nor for the announcer of Jeopardy!} represents a low discriminability (\(\kappa_j = 0.504\)) and the lowest \(\advscore_j = 0.051\) among the dataset. Although it is adversarial (\(\mu_j = 0.233\)), it fails to significantly differentiate between human and model abilities.
Similarly, for \abr{bamboogle}'s questions which were mostly \textit{reversely} adversarial, while \abr{qsr} suggested that the question is easier for humans compared to models.

\begin{table*}[!t]
    \centering
    \footnotesize
    \setlength{\tabcolsep}{4pt} 
    \renewcommand{\arraystretch}{1} 
    \begin{tabular}{p{15mm}p{44mm}p{12mm}p{9mm}p{9mm}p{17mm}p{13mm}}
    \toprule
        \textbf{Dataset} & \textbf{Question} & \textbf{Answer}  & \textbf{$\mu_j$} & \textbf{$\kappa_j$} & \textbf{Human}  & \textbf{\gpt{}-4}  \\ 
    \midrule
        \advqa{} & Who is the president of the country represented by the second letter in the acronym BRICS [...]  & Vladimir Putin  & 0.16 & 0.80 &  Putin & Russia  \\
    \midrule 
        \abr{fm2} &Henry I got married and took the throne in 1100. & True &   0.02  & 0.01 & ``True'' & False  \\
    \midrule
        \abr{trickme} &In a novel by this author, a detective wraps his arm to survive a dog attack [...]  & Durrenmatt  & 0.19 &  0.16 & ``Durrenmatt'' & Franz Kafka\\
    \midrule
        \abr{bamboogle} & Who directed the highest grossing film?  & James Cameroon & -0.02 & 0.10  & ``No idea'' &  James Cameron \\ 
    \bottomrule
    \end{tabular}
    
    \caption{\advqa{} demonstrates the most balanced properties of challenging the model and distinguishing between skills, as indicated by a positive $\mu_j$ value, which aligns with humans outperforming the models. }
    \label{tab:adv_example}
\end{table*}

\subsection{User Study}\label{userstudy}
We conducted two user studies for this paper. We recruited 1) human writers to write on the interface and 2) human respondents to answer collected \advqa{} questions and \abr{bamboogle} questions that did not have existing human responses. 
\subsection{User Study to collect questions} 
We recruited the writing team via online advertisement 
three months ahead of the human vs. computer question-answering competition. We collected 399 questions from five expert human writers (members of trivia community). 
We first display our consent form and instructions before question writers encounter the interface. They were dismissed from the study immediately if they did not pay their consent. We then inform them how their questions and prizes will be assessed; \advscore{} accurately estimates assigned criteria (e.g., adversarialness and discriminability). To make the question writing process more interesting and fun, we gamify the writing process by applying a reward system. After submitting their question sets, we calculate the \advscore{} for each writer's question set; then, we reward \$500 for those who won the first place, \$250 for second place, and \$100 for third place.

\subsection{Interface details}
\begin{figure*}[!h]
    \centering
    \includegraphics[width=0.95\textwidth]{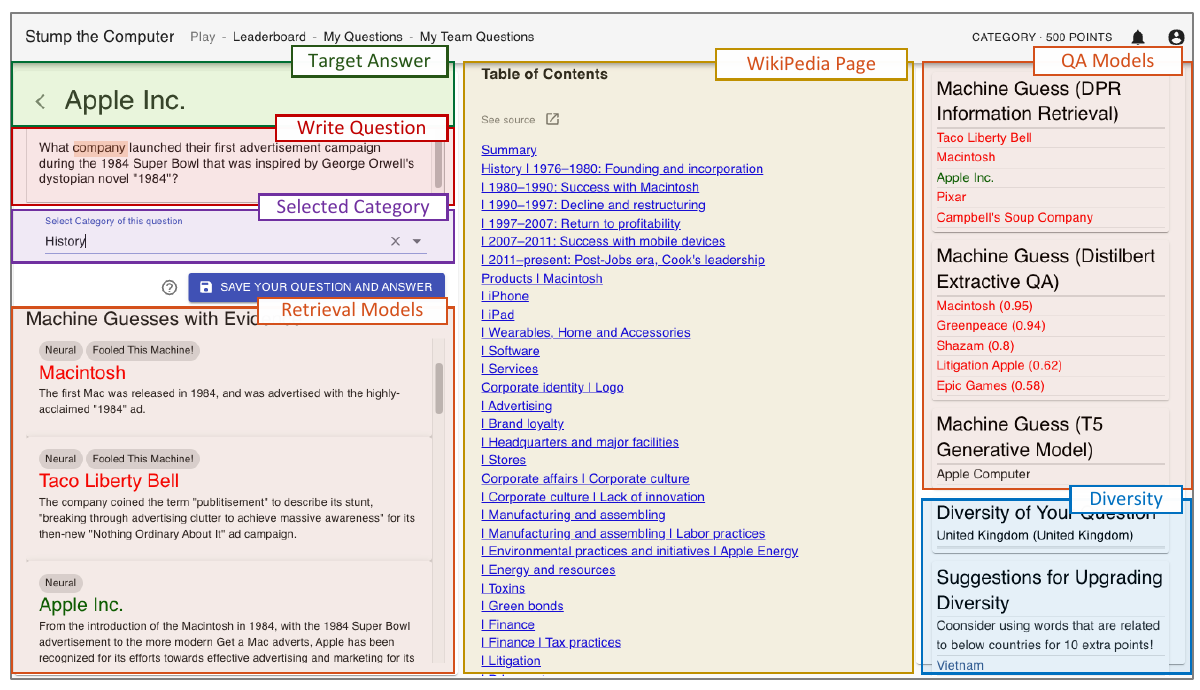}
    \caption{As the target answer to the question should be ``Apple Inc,'' the interface is updated with answers from retrieval models with the most relevant sentence and from \abr{lm}s (e.g., Distilbert, T5). Also, the highlights are updated by the input perturbation technique.} 
    \label{fig:interface}
\end{figure*}

\paragraph{Interface Screenshot}\label{interface_screenshot}
We provide an adversarial writing interface (Figure \ref{fig:interface}) as a human-AI collaborative tool for the adversarial writing competition, motivated by~\citet{you-lowd-2022-towards}'s finding that human-AI collaboration strengthens adversarial attacks. We focus on supplying the skilled-human with the real-time model interpretations, inspired by~\citet{wallace2019trick}, so that they could continuously counteract the model response and make better edits.
\paragraph{Retrieval System Details} \label{retrieval}
To ensure that the retrieval results help in obtaining up-to-date information for the writers, we created the database for Wikipedia pages and DPR training data. DPR retrieves the most relevant sentence from a database that consists of the Top 1000 popular Wikipedia pages\footnote{\url{https://pageviews.wmcloud.org/topviews/?project=en.wikipedia.org&platform=all-access&date=last-month&excludes=}} from 2021 to 2022. DPR is finetuned with the 2018 and 2021 QANTA datasets \cite{quizbowl2019}. For training, we used the questions and gold evidence as positive samples, and sentences from pages that are two hops away (pages linked by randomly selected hyperlinks in the summary section) from the question page as negative samples. 

\section{Adversarial Tactics and Question Categories} 
\subsection{Question Category Annotation} \label{app:questiontype}\label{app:topics}
We report the statistics of topic categories and adversarial tactics present in \advqa{}.
\begin{table*}[h!]
    \centering
    \footnotesize
    \setlength{\tabcolsep}{10pt} 
    \begin{tabular}{l r | l r} 
    \toprule
    \multicolumn{2}{c}{\textbf{Adversarial Tactics}} & \multicolumn{2}{c}{\textbf{Topic Categories}} \\
    \textbf{Features}  & \textbf{Count} & \textbf{Topic Category}  & \textbf{Count} \\
    \midrule
    Commonsense Knowledge       & 8   & Art            & 7   \\
    Composing Seen Clues        & 57  & Geography      & 17  \\
    Crosslingual                & 2   & History        & 33  \\
    Domain Expert Knowledge     & 10  & Lifestyle      & 11  \\
    Location Misalignment       & 10  & Literature     & 19  \\
    Logic \& Calculation        & 14  & Miscellaneous  & 31  \\
    Multi-Step Reasoning        & 50  & Music          & 13  \\
    Negation                    & 2   & Science        & 12  \\
    Novel Clues                 & 24  & Sport          & 17  \\
    Temporal Misalignment       & 5   & TV and Film    & 22  \\
    \bottomrule
    \end{tabular}
    \caption{Statistics of adversarial tactics and topics in \advqa{}}
    \label{tab:advtypes_category_counts}
\end{table*}

We ask the question writers to tag their questions with the categories below. On specific categories and examples, we encourage them to be as creative and diverse as possible when authoring the questions. In the interface, they can monitor how many questions they wrote per category. They are required to submit question sets in each of ten categories: Art, Literature,
Geography, History, Science, TV and Film, Music, Lifestyle, and Sports, Miscalleneous (Appendix~\ref{app:topics}).
\begin{table*}[t]
    \centering
    \footnotesize
    \setlength{\tabcolsep}{8pt} 
    \renewcommand{\arraystretch}{1.8}
    \begin{tabular}{p{20mm}p{120mm}}
    \Xhline{1pt}
         \textbf{Question} & \textbf{Answer} \\ 
    \Xhline{1pt}
         Art & Questions about works: Mona Lisa, Raft of the Medussa, 
     B) Questions about forms: color, contour, texture, 
     C) Questions about artists: Picasso, Monet, Leonardo da Vinci,
     D) Questions about context: Renaissance, post-modernism, expressionism, surrealism \\
     \midrule
        Literature Movement & A) Questions about works: novels (1984), plays (The Lion and the Jewel), poems (Rubaiyat), criticism (Poetics),
     B) Questions about major characters or events in literature: The Death of Anna Karenina, Noboru Wataya, the Marriage of Hippolyta and Theseus \\
         \midrule
         Literary Movement & 
A) Cross-cutting questions (appearances of Overcoats in novels),
B) Common link questions (the literary output of a country/region) \\
\midrule
        Geography & 
A) Questions about location: names of capital, state, river,
B) Questions about the place: temperature, wind flow, humidity \\
\midrule
History & A) When: When did the First World war start?,
B) Who: Who is called Napoleon of Iran?, 
C) Where: Where was the first Summer Olympics held?,
D) Which: Which is the oldest civilization in the world? \\
  \midrule
   Science & Questions about terminology: The concept of gravity was discovered by which famous physicist?, Questions about the experiment, Questions about theory: The social action theory believes that individuals are influenced by this theory.\\
\midrule
TV and Film & Quotes: What are the dying words of Charles Foster Kane in Citizen Kane?, Title: What 1927 musical was the first ``talkie''?, Plot: In The Matrix, does Neo take the blue pill or the red pill?\\
\midrule
Music & 
Singer: What singer has had a Billboard No. 1 hit in each of the last four decades?, Band: Before Bleachers and fun., Jack Antonoff fronted what band?, Title: What was Madonna's first top 10 hit?\\
\midrule
Lifestyle & Clothes: What clothing company, founded by a tennis player, has an alligator logo?, Decoration: What was the first perfume sold by Coco Chanel? \\
\midrule
Sports & Known facts: What sport is best known as the ``king of sports''?

Nationality: What is the national sport of Canada?

Sport player: The classic 1980 movie called Raging Bull is about which real-life boxer? 

Country: What country has competed the most times in the Summer Olympics yet has not won any kind of medal? \\
        \Xhline{1pt}
    \end{tabular}
    \caption{We list categories of questions along with the subcategories and corresponding examples. }
    \label{table:questiontopics}
\end{table*}

\subsection{Adversarial Tactic Annotation}\label{app:advtype}
In Table \ref{table:adversarialCategories}, we list adversarial tactics used in \advqa{} questions. We provide descriptions and examples to annotate questions with adversarial tactics (Table~\ref{table:adversarialCategories}).\footnote{Inspired by~\citet{wallace2019trick}, we add more tactics such as Location Misalignment.}
\begin{table*}[!h]
    \centering
    \footnotesize
    \setlength{\tabcolsep}{10pt} 
    \renewcommand{\arraystretch}{2.2}
    \begin{tabular}{p{50mm}p{80mm}}
    \Xhline{1pt}
    \textbf{Adversarial Type} & \textbf{Adversarial Tactics} \\ 
    \Xhline{1pt}
     Composing seen clues & Contains clues that need to be integrated for the question to be answered \\
     Logic and Calculation & Requires mathematical or logical operators \\
     Multi-Step Reasoning & Requires multiple reasoning steps between entities. For eg:  \enquote{A building dedicated to this man was the site of the ‘‘I Have A Dream’’ speech.}
    A reasoning step is required to infer : “I have a dream” speech to Lincoln Memorial to Abraham Lincoln \\
    Negation & Contains “not” or “non-” and “no” or any negation entities that may confuse the model to answer \\
    Temporal Misalignment & Contains a specific year, month, or timely event that the model is confused about or does not know.  \\
    Location Misalignment & Contains a location that the model is confused about or does not know. \\
    Commonsense Knowledge & Requires information that cannot be answered without commonsense \\
    Domain Expert Knowledge & Requires information that cannot be answered without domain expert knowledge \\
    Novel Clues & Contains information that is in the question but is not required to answer. These confuse the models. \\
    Crosslingual & Contains multilingual aspects that confuse the model. \\

    \Xhline{1pt}
    \end{tabular}
    \caption{We list adversarial tactics to determine how each question is using them to stump the models. The annotators are given the description and examples to better understand the reasons why the models may have been stumped. They are expected to tag the examples with the model prediction and question.}
    \label{table:adversarialCategories}
\end{table*}

\subsection{Annotation Examples} \label{app:annotation}
Table \ref{Tab:adv_good} shows question examples that are annotated with question and adversarial tactics. The highlights in the question correspond to either adversarial tactics or question categories that are highlighted with the same color.
\begin{table*}[!h]
    \centering
    \footnotesize
    \setlength{\tabcolsep}{3pt} 
    \renewcommand{\arraystretch}{2}
     \begin{tabular}{p{40mm}p{20mm}p{20mm}p{15mm}p{40mm}}
     \Xhline{1pt}
        \textbf{Question} & \textbf{Answer}  & \textbf{Adversarial Type}  & \textbf{Question Type}  & \textbf{Grounding} \\ 
    \Xhline{1pt}

       \colorbox{blue!15}{What is a fourth of the 5th} Bell number, often seen as an \colorbox{red!15}{unlucky} number?
        &  13/Thirteen  & \colorbox{blue!15}{Logic}\quad \colorbox{blue!15}{\& Calculation}  &  \colorbox{red!15}{Subjectivity} & ``Unlucky'' is a subjective term.\\ [1.2ex]
        
        \colorbox{red!15}{What is the famous meme} to come \colorbox{red!15}{from The Last Dance}?
        &  And I took that personally  & Composing Seen Clues &  \colorbox{red!15}{Multiple} \colorbox{red!15}{Acceptable} \colorbox{red!15}{Answers} & The meme can be referred to \textit{many} titles: ``Jordan's Cigar'', ``Jordan's Meme'', ''Laughing Jordan'', and ``Crying Jordan''\\ [1.2ex]

        What substance can cause burns in its gaseous form, lead to vomiting and sweating in high doses, and is \colorbox{blue!15}{the main component by weight}  \colorbox{blue!15}{in acid rain}?  &  Water  & \colorbox{blue!15}{Logic}\quad \colorbox{blue!15}{\& Calculation}  &  \colorbox{red!15}{Specificity} & \textit{Many substances} could cause these effects in the novel portion.\\ [1.2ex]

        Name the title character of \colorbox{red!15}{the 2024 Best Picture nominee} about a fictional conductor who Leonard Bernstein mentored.  & Lydia Tar  & Temporal Misalignment & \colorbox{red!15}{Factuality} & 2024 Best Picture Nominee \textit{cannot be factually identified} yet\\ [1.2ex]

        \colorbox{red!30}{The easternmost state in the U.S.} has more than triple its population in lakes and it is known to have \colorbox{red!15}{good salmon}, which state is it?&  Alaska  & Multihop Reasoning &  \colorbox{red!15}{Subjectivity}, \colorbox{red!30}{Specificity}  & \textit{Good salmon} is subjective, and \textit{easternmost is misleading and it requires relative position} of the author, hence non-specific.\\ [1.2ex]

    \Xhline{1pt}
    \end{tabular}
    \caption{We annotated whether each question falls into which adversarial and question type. While being adversarial; some questions lack specificity and factuality. Other questions contained subjectivity and specificity.}
    \label{Tab:adv_good}
\end{table*}

\subsection{IRT Model Details}\label{app:irt_detail}
We use a neural approach to train our 2PL IRT model, leveraging the flexibility and scalability of neural networks while maintaining the interpretability of the IRT framework. The model parameters are learned through backpropagation, with the network architecture designed to mimic the 2PL IRT structure.

\paragraph{Model Architecture}

The neural 2PL IRT model consists of three main components:
\begin{enumerate}
    \item An item embedding layer representing item difficulties ($\beta_i$) and discriminations ($\gamma_i$)
    \item A person embedding layer representing person abilities ($\theta_j$)
    \item A sigmoid output layer computing the probability of a correct response
\end{enumerate}

The total number of parameters in our model is $2N + M$, where $N$ is the number of items and $M$ is the number of subjects. This count includes $N$ difficulty parameters, $N$ discrimination parameters, and $M$ ability parameters.

\paragraph{Prior Distributions}

We incorporate prior distributions on the model parameters to enhance regularization and interpretability:

\begin{itemize}
    \item Item difficulties ($\beta_i$) and person abilities ($\theta_j$): Gaussian priors with mean 0 and variance 1
    \item Item discriminations ($\gamma_i$): Gamma prior with shape $k$ and scale $\theta$
\end{itemize}

The use of a Gamma prior for discriminations ensures positivity and allows for fine-tuning the model's sensitivity to item discrimination.

\paragraph{Training Procedure}

\begin{enumerate}
    \item Initialize network weights randomly, sampling from the respective prior distributions
    \item For each training epoch:
    \begin{enumerate}
        \item Forward pass: Compute predicted probabilities for each person-item interaction
        \item Calculate the negative log-likelihood loss
        \item Add regularization terms based on prior distributions
        \item Backpropagate the gradients and update model parameters
    \end{enumerate}
    \item Monitor validation performance and use early stopping to prevent overfitting
\end{enumerate}

We use the Adam optimizer for parameter updates due to its efficiency in treating sparse gradients and its ability to adapt the learning rate for each parameter.
\clearpage

\end{document}